\documentclass[]{spie}  

\usepackage{amsmath,amsfonts,amssymb}
\usepackage{graphicx}
\usepackage[colorlinks=true, allcolors=blue]{hyperref}

\usepackage{booktabs}       
\usepackage[capitalise]{cleveref}       
\usepackage{subcaption}     
\usepackage{svg}
\usepackage{enumitem}

\title{Enhancing Maritime Domain Awareness\\using Latent Space Representations}

\author[a]{Rikard Vinge}
\author[b]{Miguel A. Belenguer-Plomer}
\author[c]{Isabelle Wittmann}
\author[a]{Octavian Dumitru}
\author[b]{Michele Lazzarini}
\author[b]{Sergio Albani}
\author[a,d]{Conrad M Albrecht}
\affil[a]{Earth Observation Center, German Aerospace Center (DLR), Germany}
\affil[b]{European Union Satellite Centre (SatCen), 28850 Torrejon de Ardoz, Spain}
\affil[c]{IBM Research -- Europe, Switzerland}
\affil[d]{Department of Physics, Oxford University, United Kingdom}

\authorinfo{Further author information: (Send correspondence to MBP)\ MBP.: E-mail: Miguel.Belenguer@satcen.europa.eu, Telephone: +34 916786627\\
CRediT statement, \url{https://credit.niso.org}:
\textit{Conceptualization}: RV, MBP, ML, CMA;
\textit{Data curation}: RV, MBP, OD, CMA;
\textit{Funding acquisition}: SA, CMA;
\textit{Methodology}: RV, MBP, IW, CMA;
\textit{Software}: RV, MBP, IW, CMA;
\textit{Validation}: RV, MBP, OD, CMA;
\textit{Visualization}: RV, MBP;
\textit{Writing – original draft}: RV, MBP, IW, OD, CMA;
\textit{Writing – review \& editing}: RV, MBP, IW, OD, ML, SA, CMA.
\\
A major fraction of CMA's technical contribution to this work was performed while with DLR. CMA is grateful for the academic hospitality of Columbia University, New York, USA in 2025 when the majority of technical work was executed.
\\
Copyright 2026 Society of Photo-Optical Instrumentation Engineers (SPIE). One print or electronic copy may be made for personal use only. Systematic reproduction and distribution, duplication of any material in this publication for a fee or for commercial purposes, and modification of the contents of the publication are prohibited. To cite this work, please use:
\begin{quote}
    Rikard Vinge, Miguel A. Belenguer-Plomer, Isabelle Wittmann, et al., “Enhancing maritime domain awareness using latent space representations,” Proc.\ SPIE 14290, Artificial Intelligence for Security and Defence Applications IV, 1429003 (November 12, 2026). \url{https://doi.org/10.1117/12.3116838}
\end{quote}
}

\begin{document} 
\maketitle

\begin{abstract}
This work presents a novel application of Earth Observation (EO) embeddings for Maritime Domain Awareness (MDA) from the perspective of the geospatial foundation model \textit{TerraMind}. 
By operating directly on compressed latent-space representations rather than on the original Sentinel-1 imagery, this approach directly addresses the data gravity bottleneck that increasingly constrains operational EO processing, in line with the objectives of the Embed2Scale project.
In particular, linear and non-linear probing of compressed \textit{embeddings}---latent space representations of synthetic aperture radar (SAR) imagery---is used to detect, localise, and characterise vessels from Sentinel-1 GRD images. Vessel detection achieves a ROC-AUC of 0.95, while vessel localisation has a median error of around 300~m for vessels longer than 100~m, and vessel AIS type-code classification achieves 62\% accuracy for the majority classes. 
The latter, however, is highly correlated with vessel length. These results reveal both the potential and current limitations of Geospatial Foundation Models (GeoFMs) when applied to fine-grained MDA understanding in real-world scenarios.  

\noindent Additionally, a new maritime remote sensing dataset is released, pairing about 2,000 Sentinel-1 images with temporally aligned AIS signals from commercial vessels. 
Through manual inspection, each candidate sample was verified for bounding-box quality and for consistency between the AIS-reported vessel length and the observed vessel extent, cross-checking vessel identity and location against publicly available vessel records and, where available, photographs.
The dataset was made ready for use with the \textit{NeuCo-Bench} neural compression benchmark framework, publicly available on HuggingFace through \url{https://huggingface.co/datasets/embed2scale/S1-single-ship-downstream}.
\end{abstract}

\keywords{Maritime Domain Awareness, Vessel Characterisation, Radar Embedding Models, Benchmark, Foundation Model Probing, Neural Compression, Remote Sensing, Sentinel-1, AIS}

\newpage
\tableofcontents
\newpage

\section{Introduction}
\label{sec:intro}

\subsection{Motivation}
\label{sec:motivation}
The Copernicus programme generates over 22 terabytes of data per day \cite{EUSpace2026Copernicus}, a volume that keeps growing and increasingly limits how quickly Earth Observation (EO) data can be transferred, stored, and processed. This bottleneck, often referred to as \emph{data gravity}, is especially relevant for Maritime Domain Awareness (MDA), where the value of a detection depends on low-latency processing. Geospatial Foundation Models (GeoFMs) offer a potential way to ease this bottleneck by encoding EO imagery into compact latent-space representations that can be reused for downstream analysis. This study, developed within the Horizon Europe-funded Embed2Scale project, investigates whether core MDA tasks such as vessel detection, localisation, and characterisation can be performed directly from these representations, without sacrificing accuracy.

\subsection{State-of-the-Art in Maritime Domain Awareness}
\label{sec:sota}
MDA has traditionally relied on the Automatic Identification System (AIS), which broadcasts vessel identity and position but is not mandatory for all vessels and can be deactivated or spoofed \cite{alexandre2024ship}. Synthetic Aperture Radar (SAR) complements AIS by providing independent, all-weather, day-and-night observations, with Sentinel-1's open access and short revisit time making it the standard source for operational ship detection~\cite{pelich2019large}. Detection methods have evolved from statistical approaches such as Constant False Alarm Rate (CFAR) to deep learning-based object detectors, including YOLO-family architectures, which have improved both detection and classification accuracy~\cite{guan2025sar, ren2023yolo}; prior work applied YOLOv8 to an AIS-labelled Sentinel-1 dataset for this purpose~\cite{belenguer2025insights}. More recently, GeoFMs such as TerraMind \cite{jakubik2025terramind} have emerged as an alternative way to represent EO data, but their use in the maritime domain has so far been limited to feature extraction ahead of decoding or full-resolution processing. Whether MDA tasks can be performed directly within the compressed latent space itself remains largely unexplored, and is the gap this study addresses. Besides, a related limitation is the scarcity of high-quality, manually verified datasets pairing SAR imagery with AIS ground truth: existing AIS-SAR alignments are typically weakly supervised~\cite{huang2017opensarship,hammer2026ossddnewopen} and rarely cross-checked against independent vessel records, which constrains both the training and the evaluation of vessel-related EO tasks. This study also addresses this gap by releasing a manually verified maritime benchmark dataset.

\subsection{Geospatial Foundation Models}
\label{sec:geofm}
The vast scale and diversity of EO archives position them as a promising domain for foundation models~\cite{bommasani_opportunities_2021}. These are well suited to settings where high-quality labelled data are scarce but large collections of unlabelled data are available. GeoFMs are therefore pretrained through self-supervision on large EO datasets to learn general representations from unlabelled imagery~\cite{zhu2017deep,xiao2025foundation,wang2022ssl}. These representations can then support a wide range of downstream tasks using small labelled datasets and task-specific decoders~\cite{lacoste_geo-bench_2023,yue2025pangea}. A common adaptation approach is end-to-end fine-tuning, where the backbone and a task-specific decoder are optimised jointly for each task. However, this requires repeated access to the raw imagery and repeated computation through the large backbone. A more efficient paradigm treats the GeoFM as a frozen feature extractor and reuses its intermediate latent-space representations as task-agnostic embeddings~\cite{klemmer2025earth,feng2025tessera,brown2025alphaearth}. Each input scene is encoded once, after which lightweight models operate on the fixed representations. This reduces adaptation costs and enables compact representation stores, decentralised use, and similarity search at scale~\cite{vinge2026Neucobench}.

\subsection{Probing the Latent Space}
\label{sec:probing-latent-space}
When using a GeoFM as a frozen encoder, downstream performance depends on what information its latent-space representations retain and how easily it can be extracted for a given task. Probing provides a direct and low-cost way to assess this by training lightweight models on frozen representations~\cite{alain2016understanding,belinkov2022probing,vinge2026Neucobench}. Comparing linear and shallow non-linear probes further shows whether task-relevant information is directly accessible or requires a more expressive prediction model. In this work, latent-space representations generated by TerraMind from Sentinel-1 imagery~\cite{jakubik2025terramind} are probed to assess how well they capture vessel presence, type, size, and position.

\section{Datasets}
\label{sec:data}
The datasets used in this study are categorised into EO data and AIS data, as follows:
\begin{itemize}
	\item \textbf{EO}: Sentinel-1 dual-polarised (VV/VH) Ground Range Detected (GRD) backscatter coefficient images, terrain-corrected to gamma nought ($\gamma^0$) using SNAP software, were used as the primary source, downloaded from the Copernicus Data Space Ecosystem over twelve highly trafficked U.S. ports between 2020 and 2024.
	\item \textbf{AIS}: AIS records were used to derive vessel ground truth (position, type, length) at the time of each Sentinel-1 acquisition, sourced from the National Oceanic and Atmospheric Administration (NOAA).\cite{noaa_aisdata}\footnote{Additional information is available through \url{https://coast.noaa.gov/data/marinecadastre/ais/faq.pdf}.}
\end{itemize}

The Sentinel-1 imagery is not processed directly in the downstream experiments. Instead, latent-space representations are extracted using TerraMind, a GeoFM jointly developed by IBM and ESA~\cite{jakubik2025terramind}. These representations are matched with temporally aligned AIS-derived labels, as described in Section~\ref{sec:data_curation}, to construct a balanced, weakly labelled dataset. Compared with the original Sentinel-1 imagery, the latent representations reduce the data volume by a factor between 100 and 1000 depending on the model used.

This large-scale dataset is used for the vessel detection experiments described in Section~\ref{sec:results_detection}. For the vessel characterisation and localisation experiments, a smaller, manually verified subset is additionally constructed, as detailed in Section~\ref{sec:single_vessel_dataset}.

\section{Methods}
\label{sec:method}
In the following, the process of utilising the openly available data introduced in Section~\ref{sec:data} is detailed for a pipeline that curates (Sections~\ref{sec:data_curation} and~\ref{sec:single_vessel_dataset}) and embeds (Section~\ref{sec:embedding_generation}) them for (non-)linear probing (Section~\ref{sec:probing}) to detect and characterise vessels from Sentinel-1 imagery.

\subsection{AIS--Sentinel-1 Alignment}
\label{sec:data_curation}

\begin{figure}[!ht]
    \centering
    \includegraphics[width=0.3\linewidth]{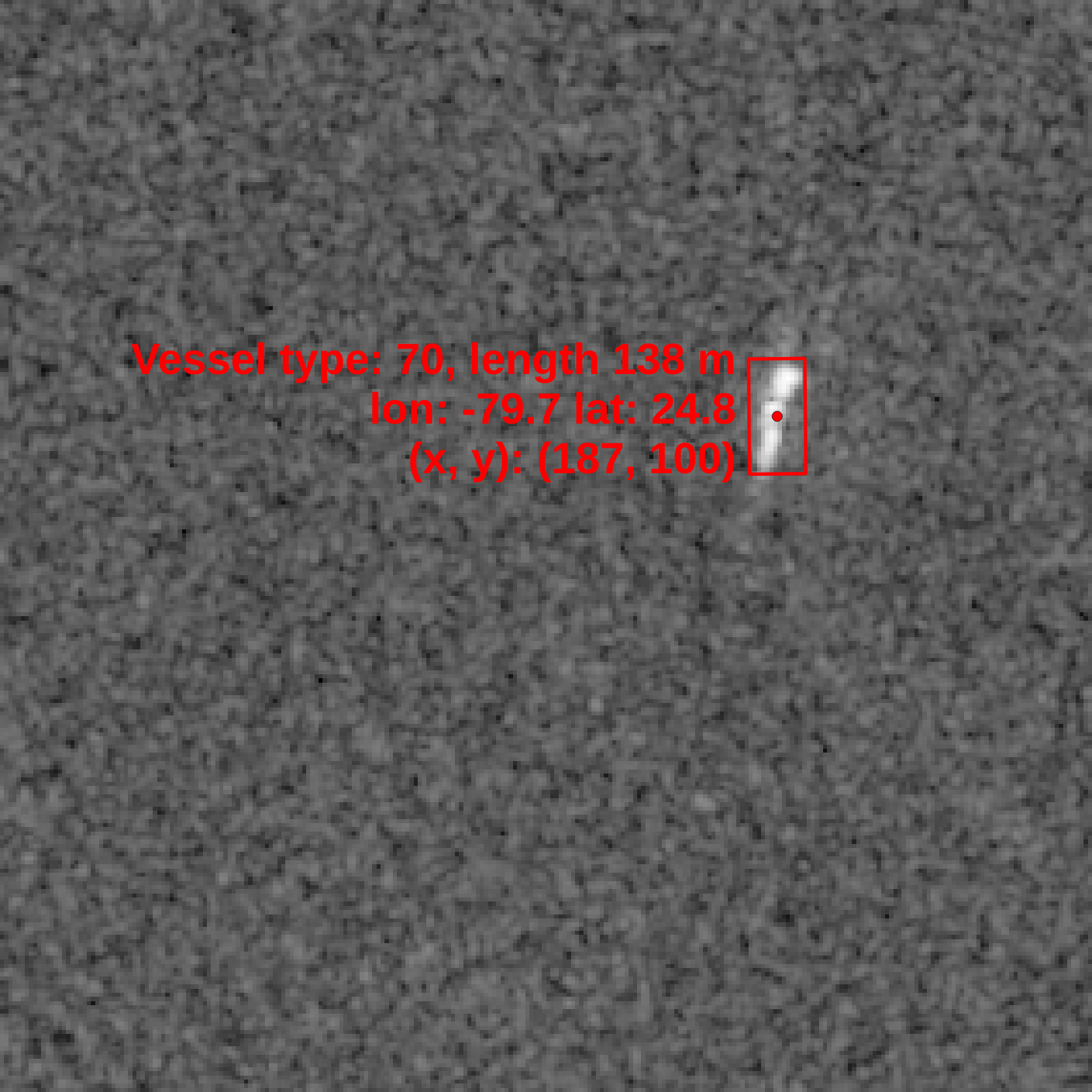}
    \caption{Example 264x264 pixel S1 greyscale image of the VV polarisation with a single vessel and a subset of metadata.}
    \label{fig:s1_metadata_example}
\end{figure}

Vessel labels were derived from AIS records and aligned with Sentinel-1 acquisitions following the same procedure as our previous study \cite{belenguer2025insights}, adapted here to operate at embedding rather than pixel level:
\begin{itemize}
	\item \textbf{Temporal matching:} only the AIS record closest in time, up to 20 minutes, to each Sentinel-1 acquisition was retained per vessel (MMSI).
	\item \textbf{Spatial refinement:} reported positions were adjusted to the nearest peak in VV/VH backscatter within a 500 m buffer, to correct for drift between the AIS report and the acquisition time. Additionally, a minimum offshore buffer of 1~km was applied to prevent misclassification with terrestrial features.
	\item \textbf{Patch-to-embedding assignment:} refined positions and AIS-derived vessel length were used to assign each vessel to its corresponding Sentinel-1 patch, and therefore to the matching TerraMind embedding.
	\item \textbf{Negative sampling:} an equal number of embeddings from patches with no AIS-confirmed vessel were sampled from the same scenes, to build a balanced dataset without biasing the negative class toward different sea or coastal conditions.
\end{itemize}
As in our prior work \cite{belenguer2025insights}, this procedure is weakly supervised: vessels not broadcasting AIS at the time of acquisition are inevitably included in the negative class, a known limitation of AIS--SAR alignment \cite{belenguer2025insights}.

\subsection{Constructing a High-Quality Downstream Dataset}
\label{sec:single_vessel_dataset}
To construct a reliable downstream dataset for ship localisation benchmarking, S1 GRD imagery was combined with vessel information derived from AIS records, followed by an extensive manual verification procedure. The goal was to retain only image crops with a single, clearly identifiable vessel and a trustworthy spatial association between the AIS record and the S1 observation.

\paragraph{Candidate Selection and Alignment with AIS Information.}
The selection started from a curated set of vessels close to the U.S. coast and its inland river streams as detailed in Section~\ref{sec:data_curation}. For each vessel, the available metadata included geographic position, timestamp, vessel dimensions, and other AIS-derived attributes. The AIS position served as the anchor for locating the corresponding vessel in the S1 imagery.

To build the downstream dataset, 3,250 vessels were first randomly sampled from the aligned S1--AIS dataset. For each sampled vessel, four image crops of size $264 \times 264$ pixels were drawn at random, with a 30 pixel (300~m) buffer such that the vessel centre point was constrained to lie within the $204 \times 204$ most central pixels of each crop. This yielded a total of 13,000 candidate crops.

\paragraph{Crop-Level Filtering and Verification.}
The number of vessels present in all 13,000 crops was manually counted. Crops containing multiple vessels were removed from further consideration. For the remaining crops with a single vessel, approximately 2,500 crops were manually inspected. The manual verification procedure focused on several aspects of quality control:
\begin{itemize}[nosep]
    \item Bounding-box quality.
    \item Consistency between the AIS-provided vessel length and the observed vessel extent in the image.
    \item Dataset-level matching status, with the following categories marked:
    \begin{itemize}[nosep]
        \item vessel location in S1 differing from the AIS data,
        \item multiple vessels visible in the S1 image,
        \item missing AIS information,
        \item vessel size mismatch,
        \item missing vessel photograph for visual inspection of ship type or missing link to the vessel location in \url{https://www.vesselfinder.com}. Crops with only this issue are kept in the final dataset.
    \end{itemize}
\end{itemize}

\paragraph{Final Dataset Assembly.}
After verification and filtering, the retained samples were compiled into a dataset consisting of 2,010 Sentinel-1 datacubes together with the corresponding metadata.

The final curated dataset prioritises label reliability over raw sample count by combining automatic AIS alignment, crop-level vessel counting, and manual inspection of candidate samples.

\paragraph{Public access to downstream dataset.}
The dataset is made publicly available\footnote{Available at \url{https://huggingface.co/datasets/embed2scale/S1-single-ship}} to extend the downstream tasks for NeuCo-Bench \cite{vinge2026Neucobench}, a benchmarking framework designed to evaluate how effectively compact, fixed-size embeddings preserve information for downstream tasks. The vessel locations and metadata originate from publicly available AIS data provided by the U.S. Marine Cadastre and cover twelve major U.S. ports. While most samples depict open water, the dataset also contains scenes intersecting land and rivers.

Each S1 sample is provided as a $264\times264$ pixel GeoTIFF containing the corresponding VV and VH polarisations of the S1 GRD product. The accompanying metadata provide the S1 acquisition time and image location as well as vessel-specific information, including IMO and MMSI identifiers, vessel type and name, dimensions, draft, cargo type, geographic coordinates, bounding-box geometry, and the pixel coordinates $(x_{\mathrm{ship}},y_{\mathrm{ship}})$ of the verified ship centre.

\paragraph{Ship labels.}
Labels compatible with NeuCo-Bench are provided in addition to the metadata. Specifically, each S1 datacube is accompanied by a structured label describing the single verified vessel contained in the image. The primary labels for the downstream localisation task are the pixel coordinates $(x_{\mathrm{ship}}, y_{\mathrm{ship}})$, which specify the centre of the vessel in the $264\times264$ pixel image, measured relative to the upper-left corner. In addition, the label contains the geographic ship position $(\mathrm{longitude},\mathrm{latitude})$ and a WKT polygon (\texttt{geometry}) describing the vessel bounding box in geographic coordinates.

The labels further retain the AIS-derived vessel information used during dataset construction and verification. This includes the vessel's IMO and MMSI identifiers, vessel type and name, length, width, draft, and cargo type. Each observation is also associated with the Unix timestamp of the S1 acquisition (\texttt{timestamp\_s1}), as well as location and time identifiers (\texttt{loc\_id} and \texttt{time\_id}). Finally, \texttt{file\_name} provides the link between a metadata record and its corresponding remote-sensing datacube. Thus, while the downstream localisation target can be represented compactly by $(x_{\mathrm{ship}},y_{\mathrm{ship}})$, the released labels retain the geographic bounding box and vessel metadata to facilitate additional analyses and alternative downstream tasks.

\noindent\textbf{Compatibility with \textit{SSL4EO-S12-downstream} and \textit{NeuCo-Bench}.}
For compatibility with the SSL4EO-S12 downstream data format, the release further contains optional colocated and coprojected Sentinel-2 L1C and L2A datacubes with the same spatial dimensions and band structure as the corresponding SSL4EO-S12 modalities. These supplementary S2 observations were deliberately selected to have 100\% cloud cover, ensuring that the ship-localisation information relevant to the downstream task must be preserved from S1 rather than inferred from optical imagery. Invisibility of the ship in the optical Sentinel-2 imagery was carefully checked by the manual ship annotation procedure. The cloudy images correspond to actual clouds at the ship's geolocation for all four seasons of the year in which the ship was captured by Sentinel-1. Within the SSL4EO-S12 data loader, the same S1 observation is replicated across all four S2 seasons. The complete release of the SSL4EO-S12-downstream task \textit{ship} has a size of approximately 14\,GB, of which the S1 imagery accounts for approximately 2.3\,GB.

\subsection{Embedding Generation with TerraMind}
\label{sec:embedding_generation}
Latent representations of the Sentinel-1 input images are obtained using TerraMind~\cite{jakubik2025terramind}, a multimodal GeoFM pretrained on the global TerraMesh corpus~\cite{blumenstiel2025terramesh}, which includes co-aligned Sentinel-1 imagery. Two complementary encoding methods are used: (i) the encoder of the modality-specific \emph{Sentinel-1 tokenizer} and (ii) the \emph{TerraMind encoder}. Both operate on the same Sentinel-1 input, allowing the value of the multimodal pretraining relative to the standalone Sentinel-1 tokenizer representation to be assessed. All encoders are kept frozen. For both encoding methods, the $264\times264$ Sentinel-1 GRD input samples are rescaled to TerraMind's pretraining input resolution of $224\times224$. Each image is then partitioned into $14\times14=196$ non-overlapping $16\times16$ patches. The channels are standardised using the TerraMesh Sentinel-1 GRD pretraining statistics.

\textbf{Sentinel-1 tokenizer embeddings}
As a unimodal baseline, representations are extracted from the encoder of TerraMind's Sentinel-1 GRD tokenizer. The tokenizer follows a quantised-autoencoder design. A ViT encoder maps each $16\times16$ pixel patch to a $5$-dimensional latent vector. Finite-scalar quantisation discretises this vector to one of $15{,}360$ possible codewords. A diffusion decoder reconstructs the image from the resulting discrete tokens. For each scene, the $196$ quantised latent vectors form a $5\times14\times14$ tensor, which is flattened before use in downstream analyses. Relative to the original input, this gives a compression factor of approximately $102\times$ when storing the quantised latent values and $1024\times$ when storing the one 16-bit token index per patch. Each token index uniquely identifies its corresponding quantised latent vector.

\textbf{TerraMind embeddings}
Unlike the unimodal tokenizer, the TerraMind encoder is pretrained jointly across multiple co-aligned EO modalities to learn cross-modal relationships. It uses a Vision Transformer that embeds each $16\times16$ pixel patch and processes the resulting sequence of patch tokens. This produces a representation of shape $N_\mathrm{patch}\times D$, where $N_\mathrm{patch}=196$ and $D$ is the encoder width. TerraMind-Base is used with $D=768$. To obtain a single embedding per image, the patch tokens are mean-pooled across the spatial dimension,
\begin{equation}
z = \frac{1}{N_\mathrm{patch}}\sum_{i=1}^{N_\mathrm{patch}} h_i
\in \mathbb{R}^{D},
\end{equation}
where $h_i$ denotes the token corresponding to patch $i$. The resulting vector $z$ provides a compact scene representation for downstream probing. This pooling is applied only to the TerraMind representations. The tokenizer representations retain their native $5\times14\times14$ spatial structure and are flattened into a $980$-dimensional vector.

\subsection{Probing}
\label{sec:probing}
To evaluate the possibility of embedding-based prediction within MDA, linear and non-linear probing are utilised. Three types of probes are employed: Random Forests \cite{breiman2001random}, Support Vector Machines \cite{cortes1995support} with polynomial kernels, and one-layer neural networks. Each probing task predicts a single target variable. Where possible, NeuCo-Bench \cite{vinge2026Neucobench} is also utilised as a standardised framework for linear and non-linear probing of EO embeddings on downstream tasks.

\section{Experiments and Discussion}
\label{sec:results}
The following experiments are organised in order of increasing task difficulty: from detecting vessel presence (Section~\ref{sec:results_detection}), to characterising vessel attributes (Section~\ref{sec:VesselCharacterization}), to precisely localising vessels within the scene (Section~\ref{sec:VesselLocalization}). This progression traces how much task-relevant information can be retained and exploited from the compressed latent space as the required granularity increases.

\subsection{Vessel Detection}
\label{sec:results_detection}
For the vessel detection task, a Random Forest (RF) probe was trained on TerraMind embeddings extracted from Sentinel-1 patches, using the AIS-derived labels described in Section~\ref{sec:data_curation}, to predict a binary vessel / no-vessel target. Hyperparameters were optimised via a randomised search (10 iterations, 3-fold cross-validation, F1-scoring) on a 30\% subsample of the training set, resulting in a 500-tree ensemble with a maximum depth of 35, a minimum of 5 samples per split, a minimum of 1 sample per leaf, \texttt{sqrt} feature sampling, and bootstrap aggregation enabled. The final model was trained on the full, class-balanced training set of 75,880 embeddings (37,940 vessel / 37,940 no-vessel) and evaluated on an independent, equally balanced validation set of 18,970 embeddings (9,485 per class). The decision threshold was tuned using Youden's J statistic (TPR $-$ FPR) on the ROC curve, to identify the operating point maximising the separation between true and false positive rates.

\begin{figure}[!ht]
	\begin{center}
		\begin{tabular}{c}
			\includegraphics[width=0.5\linewidth]{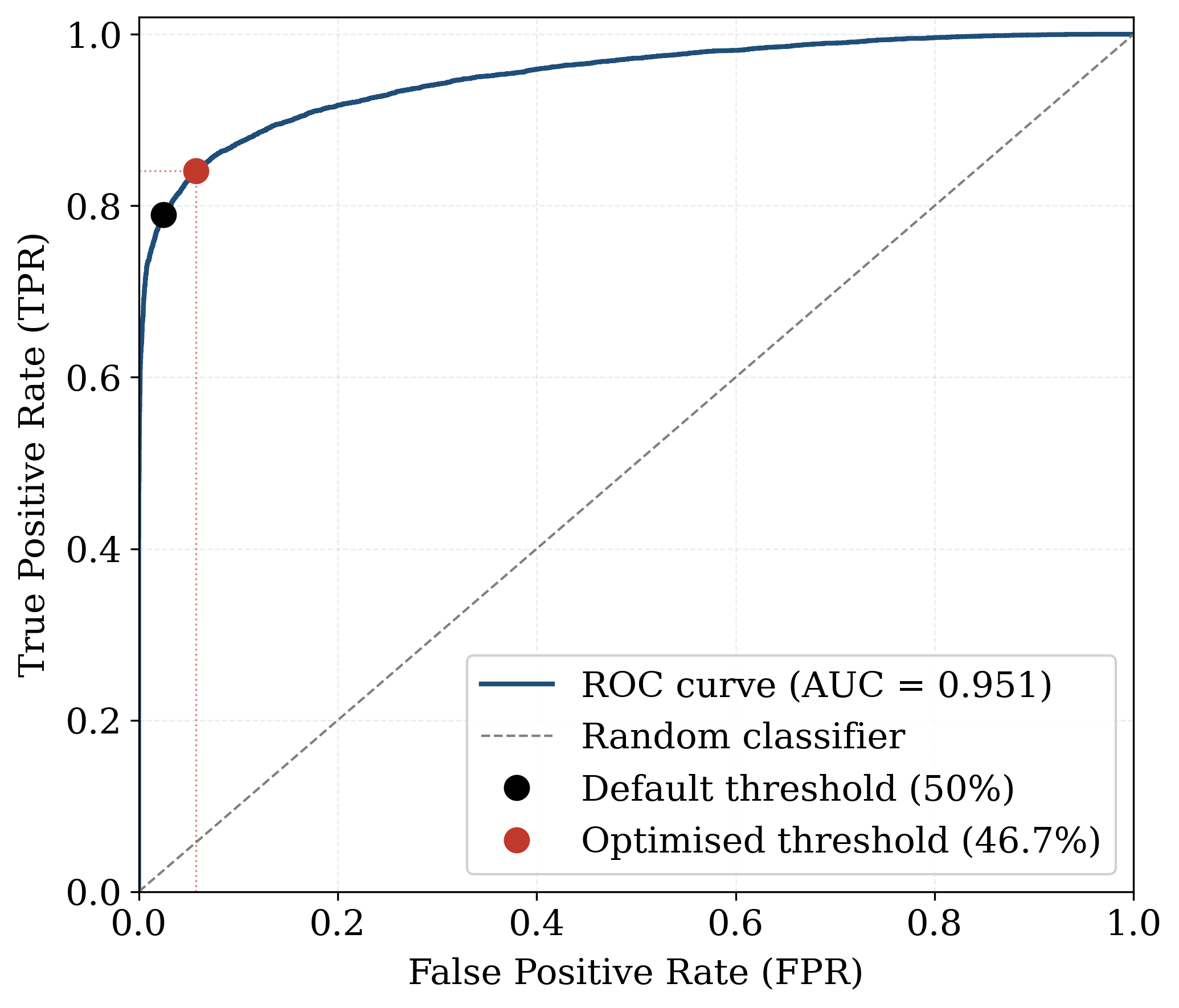}
		\end{tabular}
	\end{center}
	\caption[example]
	{\label{fig:roc_embeddings}
	ROC curve of the vessel/no-vessel classifier trained on TerraMind embeddings (ROC-AUC = 0.951), showing the default (50\%) and Youden's J-optimised (46.7\%) decision thresholds.
	}
\end{figure}

The classifier achieved a ROC-AUC of 0.951 (Figure~\ref{fig:roc_embeddings}), confirming strong separability between vessel and background embeddings in latent space. Youden's J statistic identified an optimal operating point at 46.7\%, compared to the default 50\% threshold. This shift increased recall for the Vessels class from 78.9\% to 84.1\%, at the cost of a modest rise in false positive rate from 2.5\% to 5.8\%.

At the optimised threshold, the classifier reached 94\% precision and 84\% recall for the Vessels class, and 86\% precision and 94\% recall for the No Vessels class, for an overall accuracy of 89\% (Table~\ref{tab:embeddings_classification}), with 8,935 true negatives, 550 false positives, 1,511 false negatives, and 7,974 true positives on the validation set.

\begin{table}[!ht]
	\caption{Classification report of the vessel/no-vessel classifier on TerraMind embeddings (Youden's J-optimised threshold: 46.7\%).}
	\label{tab:embeddings_classification}
	\begin{center}
		\begin{tabular}{llll}
			\hline
			\rule[-1ex]{0pt}{3.5ex}   & Precision & Recall & F1-score \\
			\hline
			\rule[-1ex]{0pt}{3.5ex}   No Vessels & 0.86 & 0.94 & 0.90 \\
			\hline
			\rule[-1ex]{0pt}{3.5ex}   Vessels & 0.94 & 0.84 & 0.89 \\
			\hline
		\end{tabular}
	\end{center}
\end{table}

These results indicate that vessel detection can be performed directly within the compressed latent space with high accuracy, without requiring access to the original Sentinel-1 imagery. The balance between precision and recall at the optimised threshold favours capturing a larger share of true vessels at a limited cost in false alarms, which is generally preferable for MDA applications where missed detections have a higher operational cost than a manageable rate of false positives.

Notably, this level of accuracy is achieved on representations that are 100 to 1{,}000 times smaller than the original Sentinel-1 imagery (Section~\ref{sec:data}), directly addressing the data-gravity bottleneck motivating this study by enabling maritime detection without transmitting or storing full-resolution SAR scenes.

\subsection{Vessel Characterisation}
\label{sec:VesselCharacterization}
Two experiments were conducted for vessel characterisation: (i) vessel length regression and (ii) vessel AIS type-code classification. To solve these tasks in embedding space, a high degree of semantic information must be preserved in the embedding. Therefore, the smaller downstream dataset was utilised for these experiments to prevent ambiguity caused by scenes containing multiple vessels.

\subsubsection{Vessel Length Prediction}
\label{sec:exp_length}
\begin{figure}[!ht]
    \centering
    \includegraphics[width=0.45\linewidth]{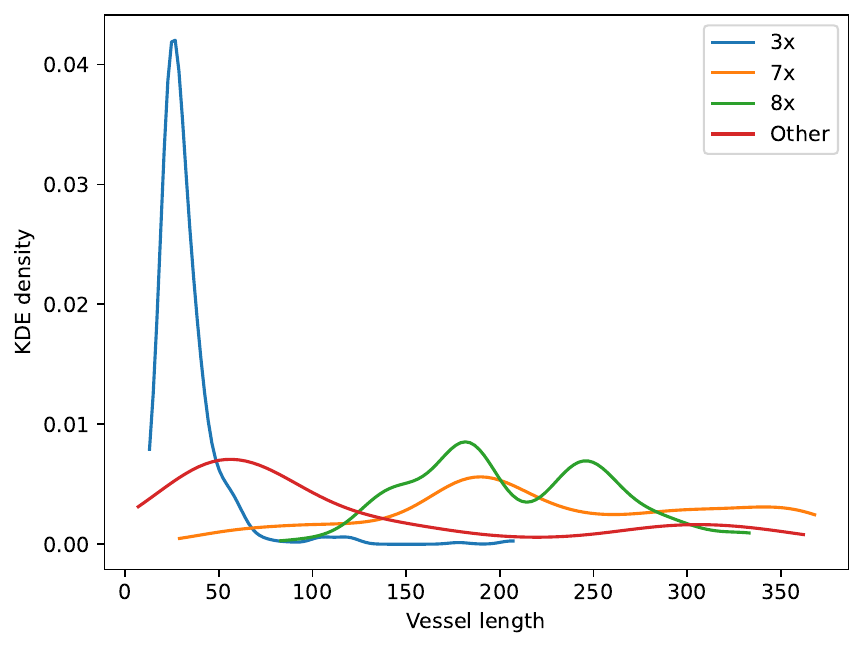}
    \caption{Kernel density estimation of the vessel length for the type categories 3x, 7x, 8x and Other -- containing all other vessel types.}
    \label{fig:vessel_length_distribution}
\end{figure}

For the vessel length task, NeuCo-Bench was utilised to evaluate linear and non-linear probes based on linear neural networks and Support Vector Machines (SVM) \cite{cortes1995support} with polynomial kernels. To assess the degree of non-linearity required for the task, the polynomial kernel SVM was employed as the primary probe model, allowing for explicit tuning of model capacity. Vessel length varies substantially depending on the vessel type, as illustrated in Figure~\ref{fig:vessel_length_distribution}. Generally, type 3x vessels are under 100~m, type 7x are broadly distributed with a heavy tail above 150~m, type 8x typically range between 100~m and 300~m, and the remaining types display a bimodal distribution with a primary peak near 60~m and a secondary peak near 300~m.

The results for both the TerraMind and TerraMind S1 tokenizer embeddings are presented in Table~\ref{tab:vessel_length}.

\begin{table}[!ht]
    \centering
    \caption{Results in $R^2$ for TerraMind and the TerraMind S1 tokenizer embeddings on predicting vessel length. A dash (-) indicates that no result above $R^2=0$ was achievable. Results presented as mean over 20 shuffle-split folds, with standard deviation over the folds in parenthesis.}
    \label{tab:vessel_length}
    \begin{tabular}{ccccc}
        \toprule
        Encoder & Linear NN & SVM 1D & SVM 2D & SVM 3D \\\hline
        S1 tokenizer & - & 0.203(0.037) & 0.251(0.044) & 0.260(0.044)\\
        TerraMind & 0.452(0.040) & 0.685(0.100) & 0.725(0.073) & 0.740(0.056)\\
        \bottomrule
    \end{tabular}
\end{table}

\begin{figure}[!ht]
    \centering
    \includegraphics[width=0.45\linewidth]{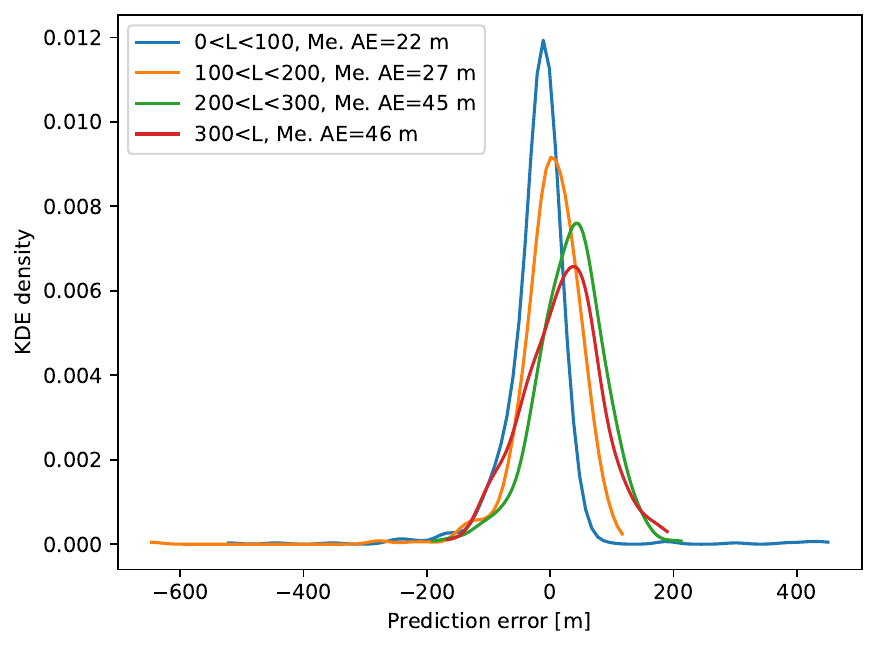}
    \includegraphics[width=0.45\linewidth]{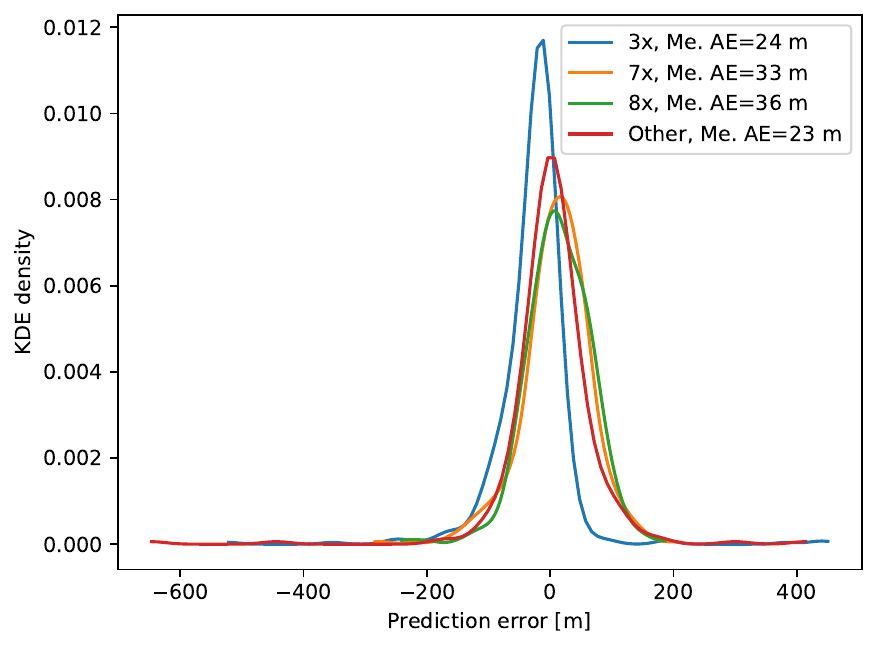}
    \caption{Kernel density estimation of the vessel length prediction error for the 2D polynomial SVM. (Left) with respect to vessel length and (right) with respect to vessel type. Median absolute error provided for each group.}
    \label{fig:vessel_length_results}
\end{figure}

The 2D and 3D SVM probes successfully predict vessel length. In Figure~\ref{fig:vessel_length_results}, where errors are stratified by vessel length and type, absolute prediction errors increase for longer vessels and those belonging to categories 7x and 8x (which skew larger; Section~\ref{sec:exp_length}). However, relative error decreases with length: for vessels under 100~m, the median absolute error is 60\% of vessel length; for 100--200~m vessels, it falls to 17\%; for 200--300~m vessels, it is 18\%; and for vessels over 300~m, the median error drops to 14\% of total length.

\subsubsection{Vessel Type Classification}
\label{sec:exp_type}
To predict AIS vessel types (Table~\ref{tab:vessel_type_distr}), a Random Forest classifier was trained using FLAML \cite{Wang_FLAML_A_Fast_2021} on single-vessel patches. Because the curated downstream dataset is highly imbalanced, primary classification results are reported across the three majority classes: 3x, 7x, and 8x. When incorporating a fourth ``Other'' category or attempting 10-class classification, severe class imbalance caused the model to collapse minority predictions into the three majority classes.

\begin{table}[!ht]
    \centering
    \caption{Vessel type distribution in the downstream dataset. In parentheses are the vessel type classes used in the vessel type prediction experiments.}
    \label{tab:vessel_type_distr}
    \begin{tabular}{llc}
        \toprule
        AIS type code & Category & Number of vessels \\\hline
        0 & Not available & 2\\
        1 - 19 & Reserved & 3\\
        20 - 29 & Wing-in-Ground & 2\\
        30 - 39 (3x) & Fishing, towing, diving & 533\\
        40 - 49 & High speed craft & 1\\
        50 - 59 & Special craft & 41\\
        60 - 69 & Passenger & 65\\
        70 - 79 (7x) & Cargo & 776\\
        80 - 89 (8x) & Tanker & 411\\
        90 - 99 & Other & 176\\
        \bottomrule
    \end{tabular}
\end{table}

\begin{figure}[!ht]
    \centering
    \begin{subfigure}{0.47\textwidth}
        \includegraphics[width=0.49\textwidth]{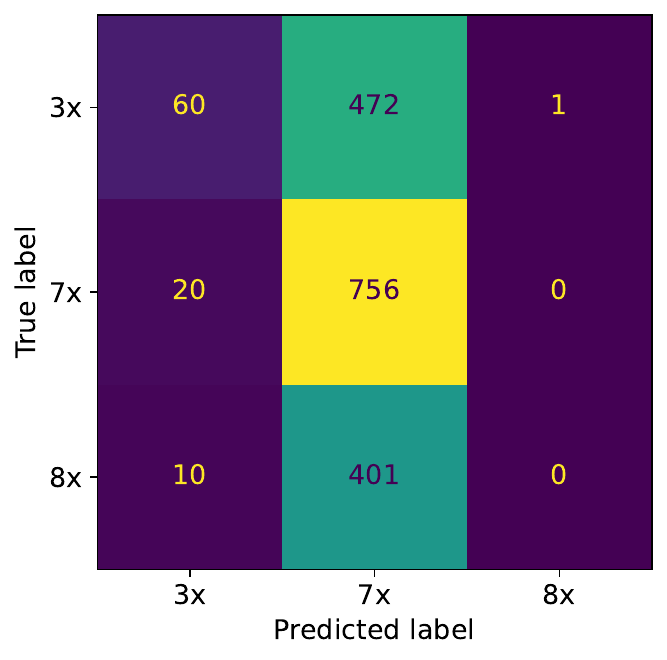}
        \includegraphics[width=0.49\textwidth]{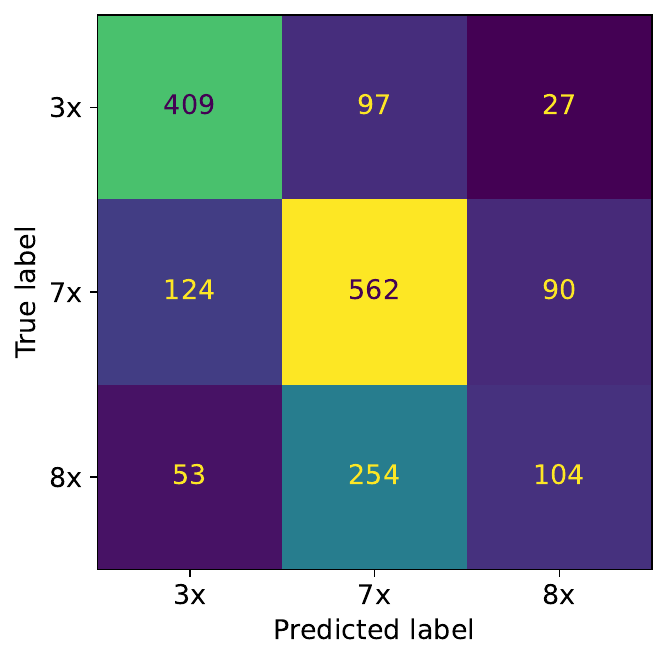}
        \caption{Embeddings from (left) TerraMind S1 tokenizer and (right) TerraMind, all vessels.}
        \label{fig:vessel_type_confmat_all}
    \end{subfigure}
    \hfill
    \begin{subfigure}{0.47\textwidth}
        \includegraphics[width=0.49\textwidth]{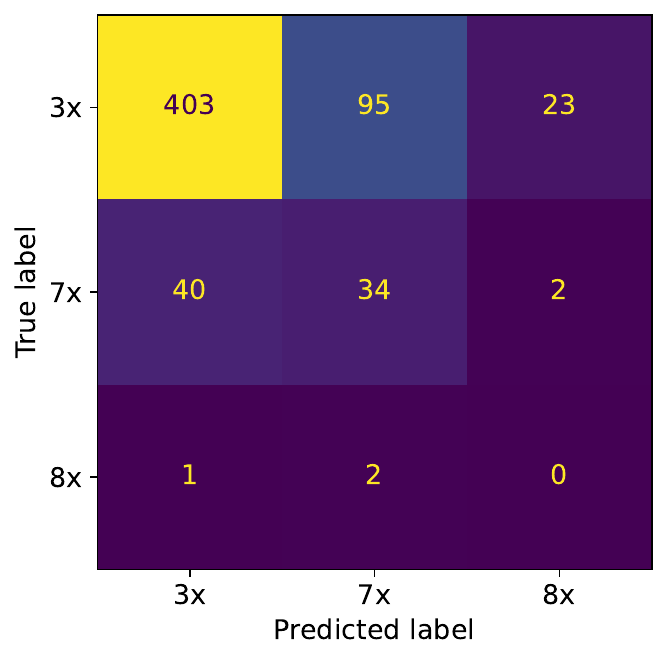}
        \includegraphics[width=0.49\textwidth]{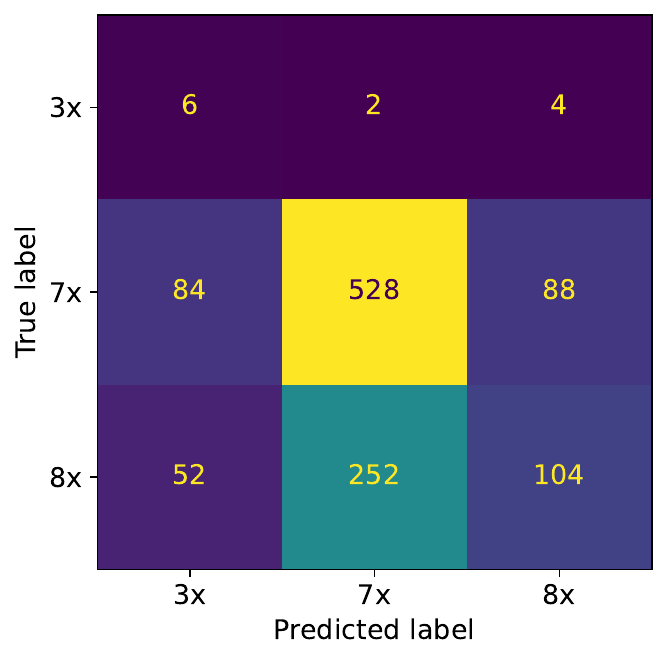}
        \caption{TerraMind embeddings, (left) vessels shorter than 100~m, (right) vessels longer than 100~m.}
        \label{fig:vessel_type_confmat_wrt_length}
    \end{subfigure}
    \vspace{6pt}
    \caption{Confusion matrices for vessel AIS type code classification on (a) TerraMind and its S1 tokenizer embeddings and (b) TerraMind embeddings when separated on vessels shorter and longer than 100 m. See Table~\ref{tab:vessel_type_distr} for label designations.}
    \label{fig:vessel_type_confmat}
\end{figure}

The results for TerraMind and the S1 tokenizer of TerraMind are presented in Table~\ref{tab:vessel_type_results}, with confusion matrices shown in Figure~\ref{fig:vessel_type_confmat_all}.

\begin{table}[!ht]
    \centering
    \caption{10-fold cross-validation results for vessel type prediction between the 3x, 7x, and 8x classes presented as mean over folds and standard deviation over folds in parenthesis. The indication (W) for F1, precision and recall denotes a weighted metric.}
    \label{tab:vessel_type_results}
    \begin{tabular}{lcccccc}
        \toprule
        Encoder & Accuracy & F1 Macro & F1 Micro & F1 (W) & Precision (W) & Recall (W) \\\hline
        S1 tokenizer & 0.474(0.041) & 0.273(0.022) & 0.474(0.041) & 0.344(0.044) & 0.436(0.054) & 0.474(0.041) \\
        TerraMind & 0.624(0.036) & 0.572(0.028) & 0.624(0.036) & 0.604(0.036) & 0.612(0.031) & 0.624(0.036) \\
    \end{tabular}
\end{table}

Generally, the TerraMind embeddings perform significantly better compared to the S1 tokenizer. The reason is mainly that the classifier trained on the S1 tokenizer embeddings over-predicts to type 7x. The TerraMind embeddings are considerably more balanced but also overassign samples to the largest of the classes. When splitting the results based on vessel length (Figure~\ref{fig:vessel_type_confmat_wrt_length}), it was found that the classifier based on TerraMind embeddings in fact classifies roughly given the threshold ``vessel length greater or smaller than 100 m''. Given the different vessel length distributions for the classes (Section~\ref{sec:exp_length}), it cannot be concluded that vessel type identification is achievable further than based on the vessel length.

\subsection{Vessel Localisation}
\label{sec:VesselLocalization}
The final set of experiments aims to investigate the possibility of localising vessels directly from the embeddings. It relies on two steps: (i) a quadrant prediction task in which a binary classifier was trained to predict whether the vessel is in the upper or lower half of the image and another to predict whether it is on the left or right side; and (ii) predicting the vessel pixel location using the centre point of the bounding box as the target. The experiments were performed using NeuCo-Bench with the same setup as for the vessel length prediction.

\subsubsection{Quadrant Localisation}
\label{sec:exp_quadrant}
The quadrant prediction experiment constitutes two almost perfectly balanced binary classification problems, due to the uniform location distribution of the vessels within the images.

The results for the quadrant localisation are given in Table~\ref{tab:exp_quadrant_localization}. As with the vessel length prediction experiments, TerraMind embeddings and a 2D or 3D kernel SVM probe generally perform well on this task, achieving F1 scores of 0.76 and accuracy scores of the same magnitude. The S1 tokenizer embeddings slightly outperform TerraMind when using the linear network but perform worse on the SVM probes.

\begin{table}[!ht]
    \centering
    \caption{Results in F1 score for TerraMind and the TerraMind S1 tokenizer embeddings on predicting the image quadrant containing the vessel. Results presented as mean over 20 shuffle-split folds, with standard deviation over the folds in parenthesis.}
    \label{tab:exp_quadrant_localization}
    \begin{tabular}{cccccc}
        \toprule
        Encoder & Direction & Linear NN & SVM 1D & SVM 2D & SVM 3D \\\hline
        S1 tokenizer & Horizontal & 0.683(0.029) & 0.677(0.027) & 0.708(0.024) & 0.711(0.025)\\
        S1 tokenizer & Vertical & 0.682(0.037) & 0.664(0.038) & 0.679(0.033) & 0.673(0.028)\\
        TerraMind & Horizontal & 0.671(0.032) & 0.737(0.025) & 0.760(0.027) & 0.756(0.027)\\
        TerraMind & Vertical & 0.675(0.037) & 0.752(0.030) & 0.787(0.028) & 0.785(0.026)\\
        \bottomrule
    \end{tabular}
\end{table}

When the vessels are separated by length and type, it was found that short vessels have lower scores, with F1 scores closer to 0.6 for the 2D and 3D SVM models, while the F1-score performance is around 0.8--0.9 for vessels of lengths greater than 100 m. Similarly, the longer 7x and 8x types have F1-scores around 0.85, while the shorter 3x vessels are closer to 0.6 in F1-score. No significant performance dependency on the closeness of the vessel to the classification boundary, i.e., the vertical or horizontal line through the centre pixel, was found.

\subsubsection{Pixel Localisation}
For pixel localisation, the centre point of the bounding box was used as a proxy for the vessel location. Two regression models were trained, the first to predict the horizontal pixel index of the bounding box centre point, and the second to predict the vertical pixel index. The pixel indices were normalised by subtracting the half-width/height of the image and dividing by the full width/height, yielding a target range in approximately [-0.39, 0.39] due to the 30 pixel buffer from the vessel centre point to the image boundary (Section~\ref{sec:single_vessel_dataset}). As with the vessel length and quadrant localisation experiments, NeuCo-Bench was utilised for this experiment.

The results for the pixel localisation are given in Table~\ref{tab:exp_pixel_localization}.

\begin{table}[!ht]
    \centering
    \caption{Results in $R^2$ for TerraMind and the TerraMind S1 tokenizer embeddings on predicting pixel location of the vessel centre. A dash (-) indicates that no result above $R^2=0$ was achievable. Results presented as mean over 20 shuffle-split folds, with standard deviation over the folds in parenthesis.}
    \label{tab:exp_pixel_localization}
    \begin{tabular}{cccccc}
        \toprule
        Encoder & Direction & Linear NN & SVM 1D & SVM 2D & SVM 3D \\\hline
        S1 tokenizer & Horizontal & - & 0.156(0.043) & 0.183(0.033) & 0.183(0.033)\\
        S1 tokenizer & Vertical & - & 0.142(0.036) & 0.140(0.029) & 0.138(0.027)\\
        TerraMind & Horizontal & 0.025(0.026) & 0.208(0.106) & 0.207(0.078) & 0.207(0.079)\\
        TerraMind & Vertical & 0.065(0.039) & 0.275(0.085) & 0.332(0.067) & 0.328(0.065)\\
        \bottomrule
    \end{tabular}
\end{table}

\begin{figure}[!ht]
    \centering
    \begin{subfigure}{0.4\textwidth}
        \includegraphics[width=\textwidth]{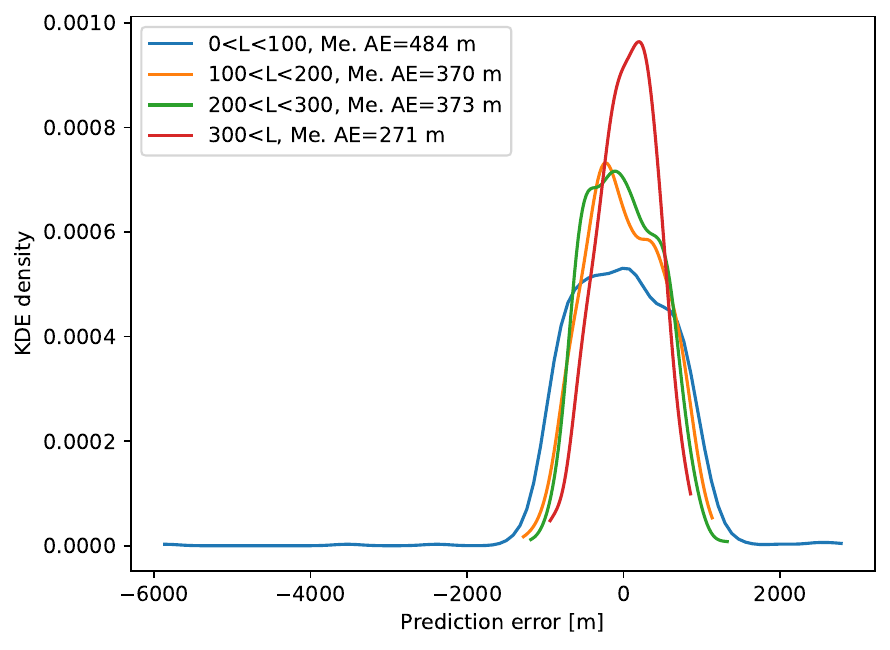}
        \caption{Horizontal localisation wrt. vessel length.}
        \label{fig:vessel_localization_px_hl}
    \end{subfigure}
    \begin{subfigure}{0.4\textwidth}
        \includegraphics[width=\textwidth]{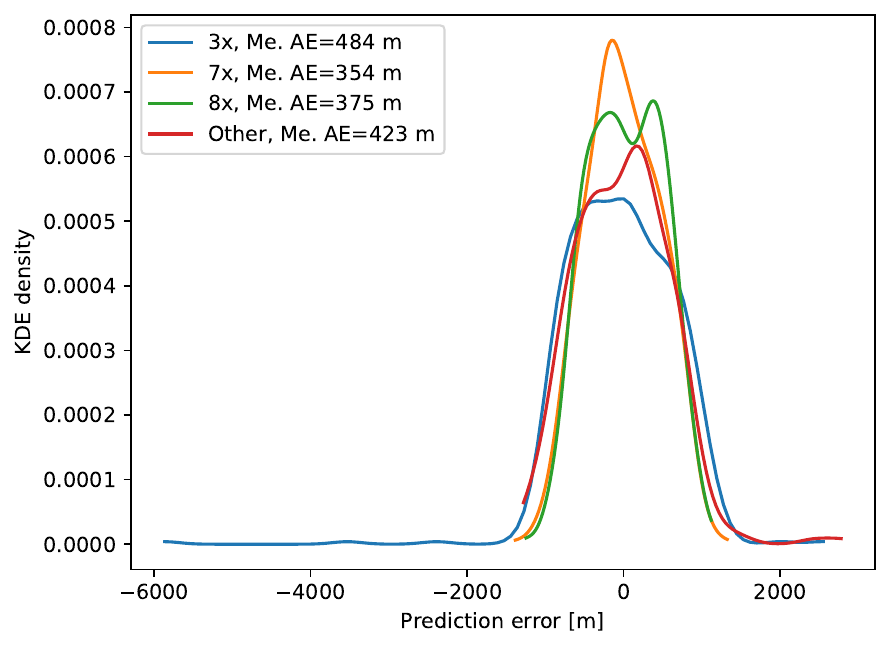}
        \caption{Horizontal localisation wrt. vessel type.}
        \label{fig:vessel_localization_px_ht}
    \end{subfigure}
    \begin{subfigure}{0.4\textwidth}
        \includegraphics[width=\textwidth]{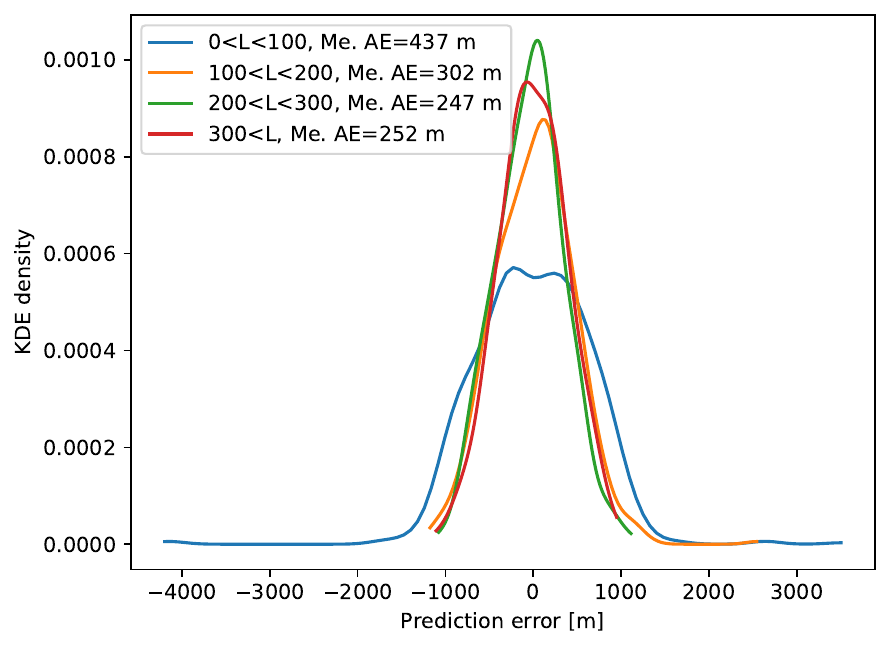}
        \caption{Vertical localisation wrt. vessel length.}
        \label{fig:vessel_localization_px_vl}
    \end{subfigure}
    \begin{subfigure}{0.4\textwidth}
        \includegraphics[width=\textwidth]{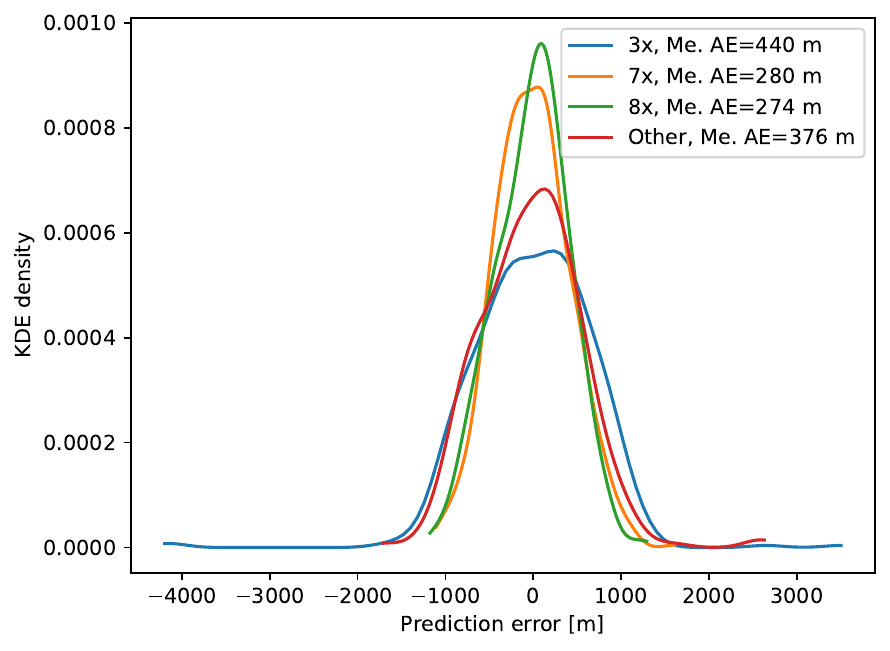}
        \caption{Vertical localisation wrt. vessel type.}
        \label{fig:vessel_localization_px_vt}
    \end{subfigure}
    \caption{Kernel density estimation of the vessel location prediction error for the 2D polynomial SVM. Median absolute error provided for each group.}
    \label{fig:vessel_localization_px}
\end{figure}

Generally, the pixel index prediction task is difficult for all probes. It was found that the horizontal and vertical prediction performances differ slightly, although they overlap within one standard deviation. When separated by vessel length and type (Figure~\ref{fig:vessel_localization_px}), it was found that vessels shorter than 100 m have median errors of 440--480 m, while longer vessels can be predicted down to within 250--370 m, about 10--15\% of the image, or within about two $16\times16$ pixel patches.

Although the $R^2$ results are low, such localisation performances can still be valuable as an indication of the location of the vessel at a finer resolution than the quadrant.

\section{Conclusions}
\label{sec:conclusions}

This work demonstrates that TerraMind embeddings retain features relevant to Maritime Domain Awareness, supporting vessel detection, characterisation, and localisation directly in latent space. The representations achieved compression factors of approximately 100 to 200, rising to over 1{,}000 with embedding-space quantisation of the TerraMind Sentinel-1 tokenizer. These results support the objectives of Embed2Scale by showing that maritime information can be extracted from compressed representations. For the tasks evaluated, this approach can reduce the volume of full-resolution SAR imagery that needs to be transmitted, stored, and processed, helping to alleviate the data gravity bottleneck in EO-based maritime surveillance. This work also releases a benchmark of 2{,}010 manually verified single-vessel Sentinel-1 crops with cross-checked AIS metadata, addressing the scarcity of labelled data for vessel-related EO tasks and enabling reproducible evaluation through NeuCo-Bench.

Vessel presence and length were the most reliably inferred properties. A Random Forest probe achieved a ROC-AUC of 0.951 for vessel presence, with 94\% precision and 84\% recall at the threshold that maximised Youden's $J$. Vessel length estimation reached an $R^2$ of up to 0.74, with lower relative errors for longer vessels. AIS type-code characterisation achieved 62\% accuracy for the majority classes (AIS type codes in the 30s, 70s, and 80s), although the predicted types were strongly associated with vessel length. In the tasks for which both representations were evaluated, TerraMind embeddings outperformed the TerraMind Sentinel-1 tokenizer representations, with the largest advantage in characterisation.

Localisation performance depended on the spatial precision required. Quadrant prediction achieved F1 scores of 0.76--0.79, while pixel-level localisation yielded median errors of 250--480~m, depending on vessel size. Performance was generally poorer for smaller vessels across the tasks evaluated. This pattern is consistent with class imbalance and the limited spatial detail available at a Sentinel-1 pixel spacing of 10~m, particularly when the embedding process divides the imagery into $16 \times 16$-pixel ($160 \times 160$~m) patches and averages their representations. The reported metrics should also be interpreted in light of the AIS-based reference data, which exclude vessels that were not broadcasting AIS at the time of acquisition.

Taken together, the results show that compressed GeoFM embeddings preserve information for vessel presence detection and length estimation, while AIS type-code characterisation and precise localisation remain more challenging. Establishing the operational value of latent-space processing will require a direct comparison with conventional image-based pipelines. Future work will pursue this comparison, including object detection on reconstructed imagery, and will investigate higher-capacity probes for localisation, extend characterisation to the full 10-class AIS taxonomy, and evaluate additional GeoFMs beyond TerraMind.

\section*{ACKNOWLEDGMENTS AND DISCLAIMER}    
This work was funded by \textit{Embed2Scale} which is co-funded by the EU Horizon Europe programme under Grant Agreement No.~\texttt{101131841}. Additional funding for this project has been provided by the Swiss State Secretariat for Education, Research and Innovation (SERI) and UK Research and Innovation (UKRI). 

The views expressed are solely those of the author(s) and not of the European Union Satellite Centre (SatCen), which accepts no responsibility for their use. The authors declare no conflict of interest.

\bibliography{references}

@misc{EUSpace2026Copernicus,
  author       = {{European Union Agency for the Space Programme}},
  title        = {Earth Observation: {Copernicus} and {EOGS}},
  year         = {2026},
  note         = {\url{https://eu-space.europa.eu/programmes/earth-observation-copernicus-and-eogs}. Accessed: 2026-08-19}
}

@inproceedings{jakubik2025terramind,
  title={{Terramind: Large-scale generative multimodality for earth observation}},
  author={Jakubik, Johannes and Yang, Felix and Blumenstiel, Benedikt and Scheurer, Erik and Sedona, Rocco and Maurogiovanni, Stefano and Bosmans, Jente and Dionelis, Nikolaos and Marsocci, Valerio and Kopp, Niklas and others},
  booktitle={2025 IEEE/CVF International Conference on Computer Vision (ICCV)},
  pages={7383--7394},
  year={2025},
  organization={IEEE}
}

@inproceedings{belenguer2025insights,
  title={{Insights into deep learning-based vessel detection and characterization using SAR and AIS data}},
  author={Belenguer-Plomer, Miguel A and Lazzarini, Michele and Barrilero, Omar and Saame{\~n}o, Paula and Albani, Sergio},
  booktitle={Artificial Intelligence for Security and Defence Applications III},
  volume={13679},
  pages={206--210},
  year={2025},
  organization={SPIE}
}

@article{guan2025sar,
	title={{SAR Small Ship Detection Based on Enhanced YOLO Network}},
	author={Guan, Tianyue and Chang, Sheng and Wang, Chunle and Jia, Xiaoxue},
	journal={Remote Sensing},
	volume={17},
	number={5},
	pages={839},
	year={2025},
	publisher={MDPI}
}

@article{ren2023yolo,
	title={{YOLO-Lite: An efficient lightweight network for SAR ship detection}},
	author={Ren, Xiaozhen and Bai, Yanwen and Liu, Gang and Zhang, Ping},
	journal={Remote Sensing},
	volume={15},
	number={15},
	pages={3771},
	year={2023},
	publisher={MDPI}
}

@article{alexandre2024ship,
	title={{Ship detection with SAR C-Band satellite images: A systematic review}},
	author={Alexandre, Cyprien and Devillers, Rodolphe and Mouillot, David and Seguin, Raphael and Catry, Thibault},
	journal={IEEE Journal of Selected Topics in Applied Earth Observations and Remote Sensing},
    volume={17},
    pages={14353--14367},
    year={2024},
    publisher={IEEE}
}

@article{pelich2019large,
	title={{Large-scale automatic vessel monitoring based on dual-polarization Sentinel-1 and AIS data}},
	author={Pelich, Ramona and Chini, Marco and Hostache, Renaud and Matgen, Patrick and Lopez-Martinez, Carlos and Nuevo, Miguel and Ries, Philippe and Eiden, Gerd},
	journal={Remote Sensing},
	volume={11},
	number={9},
	pages={1078},
	year={2019},
	publisher={MDPI}
}

@misc{hammer2026ossddnewopen,
      title={{OSSDD - a New Open Dataset for Sentinel-1 Ship Detection}}, 
      author={Horst Hammer and Sylvia Hochstuhl and Antje Thiele and Tobias Brosch and Padraig Davidson and Tim Remiger and Michael Teutsch},
      year={2026},
      eprint={2608.01963},
      archivePrefix={arXiv},
      primaryClass={cs.CV},
      note={arXiv:2608.01963}, 
}

@article{huang2017opensarship,
  title={{OpenSARShip: A dataset dedicated to Sentinel-1 ship interpretation}},
  author={Huang, Lanqing and Liu, Bin and Li, Boying and Guo, Weiwei and Yu, Wenhao and Zhang, Zenghui and Yu, Wenxian},
  journal={IEEE Journal of Selected Topics in Applied Earth Observations and Remote Sensing},
  volume={11},
  number={1},
  pages={195--208},
  year={2017},
  publisher={IEEE}
}

@InProceedings{vinge2026Neucobench,
    author={Vinge, Rikard and Wittmann, Isabelle and Schneider, Jannik and Marszalek, Michael and Gilch, Luis and Brunschwiler, Thomas and Albrecht, Conrad M},
    title={{NeuCo-Bench: A Novel Benchmark Framework for Neural Embeddings in Earth Observation}},
    booktitle={Proceedings of the IEEE/CVF Conference on Computer Vision and Pattern Recognition (CVPR) Workshops},
    month={June},
    year={2026},
    pages={8063-8074}
}

@article{breiman2001random,
  title={Random forests},
  author={Breiman, Leo},
  journal={Machine learning},
  volume={45},
  number={1},
  pages={5--32},
  year={2001},
  publisher={Springer}
}

@article{cortes1995support,
  title={Support-vector networks},
  author={Cortes, Corinna and Vapnik, Vladimir},
  journal={Machine learning},
  volume={20},
  number={3},
  pages={273--297},
  year={1995},
  publisher={Springer}
}

@article{Wang_FLAML_A_Fast_2021,
  title={{FLAML: A fast and lightweight automl library}},
  author={Wang, Chi and Wu, Qingyun and Weimer, Markus and Zhu, Erkang},
  journal={Proceedings of machine learning and systems},
  volume={3},
  pages={434--447},
  year={2021}
}

@misc{noaa_aisdata,
    author = {{NOAA Office for Coastal Management}},
    title = {Vessel Traffic Broadcast Points [dataset]},
    howpublished = {Marine Cadastre},
    year = {2020-2024},
    note = {\url{https://marinecadastre.gov}}
}

@misc{bommasani_opportunities_2021,
      title={{On the Opportunities and Risks of Foundation Models}}, 
      author={Bommasani, Rishi and Hudson, Drew A and Adeli, Ehsan and Altman, Russ and Arora, Simran and von Arx, Sydney and Bernstein, Michael S and Bohg, Jeannette and Bosselut, Antoine and Brunskill, Emma and others},
      year={2022},
      eprint={2108.07258},
      archivePrefix={arXiv},
      primaryClass={cs.LG},
      note={arXiv:2108.07258}, 
}

@article{lacoste_geo-bench_2023,
      title={{GEO-Bench: Toward Foundation Models for Earth Monitoring}},
      author={Alexandre Lacoste and Nils Lehmann and Pau Rodriguez and Evan Sherwin and Hannah Kerner and Bj{\"o}rn L{\"u}tjens and Jeremy Irvin and David Dao and Hamed Alemohammad and Alexandre Drouin and others},
      journal={Advances in Neural Information Processing Systems},
      volume={36},
      pages={51080--51093},
      year={2023}
}

@inproceedings{yue2025pangea,
      title={{Pangea: A Fully Open Multilingual Multimodal LLM for 39 Languages}},
      author={Xiang Yue and Yueqi Song and Akari Asai and Seungone Kim and Jean de Dieu Nyandwi and Simran Khanuja and Anjali Kantharuban and Lintang Sutawika and Sathyanarayanan Ramamoorthy and Graham Neubig},
      booktitle={International Conference on Learning Representations},
      year={2025},
      url={https://openreview.net/forum?id=a3g2l4yEys}
}

@article{zhu2017deep,
      title={{Deep Learning in Remote Sensing: A Review}},
      author={Xiao Xiang Zhu and Devis Tuia and Lichao Mou and Gui-Song Xia and Liangpei Zhang and Feng Xu and Friedrich Fraundorfer},
      journal={IEEE Geoscience and Remote Sensing Magazine},
      volume={5},
      number={4},
      pages={8--36},
      year={2017},
      doi={10.1109/MGRS.2017.2762307}
}

@article{wang2022ssl,
      title={{Self-Supervised Learning in Remote Sensing: A Review}},
      author={Yi Wang and Conrad M. Albrecht and Nassim Ait Ali Braham and Lichao Mou and Xiao Xiang Zhu},
      journal={IEEE Geoscience and Remote Sensing Magazine},
      volume={11},
      number={3},
      pages={22--51},
      year={2023},
      doi={10.1109/MGRS.2023.3247548}
}

@article{xiao2025foundation,
      title={{Foundation Models for Remote Sensing and Earth Observation: A Survey}},
      author={Aoran Xiao and Weihao Xuan and Junjue Wang and Jiaxing Huang and Dacheng Tao and Shijian Lu and Naoto Yokoya},
      journal={IEEE Geoscience and Remote Sensing Magazine},
      volume={13},
      number={4},
      pages={297--324},
      year={2025},
      doi={10.1109/MGRS.2025.3576766}
}

@article{klemmer2025earth,
  author={Klemmer, Konstantin and Rolf, Esther and Rußwurm, Marc and Camps-Valls, Gustau and Czerkawski, Mikolaj and Ermon, Stefano and Francis, Alistair and Jacobs, Nathan and Kerner, Hannah and Mackey, Lester and Mai, Gengchen and Aodha, Oisin Mac and Reichstein, Markus and Robinson, Caleb and Rolnick, David and Shelhamer, Evan and Sitzmann, Vincent and Tuia, Devis and Zhu, Xiaoxiang},
  journal={IEEE Geoscience and Remote Sensing Magazine}, 
  title={Earth Embeddings: Toward artificial intelligence-centric representations of our planet}, 
  year={2026},
  volume={},
  number={},
  pages={2-15},
  doi={10.1109/MGRS.2026.3710416}
}

@misc{feng2025tessera,
      title={{TESSERA: Temporal Embeddings of Surface Spectra for Earth Representation and Analysis}},
      author={Zhengpeng Feng and Clement Atzberger and Sadiq Jaffer and Jovana Knezevic and Silja Sormunen and Robin Young and Madeline C. Lisaius and Markus Immitzer and Toby Jackson and James Ball and David A. Coomes and Anil Madhavapeddy and Andrew Blake and Srinivasan Keshav},
      year={2025},
      eprint={2506.20380},
      archivePrefix={arXiv},
      primaryClass={cs.LG},
      note={arXiv:2506.20380}
}

@misc{brown2025alphaearth,
      title={{AlphaEarth Foundations: An Embedding Field Model for Accurate and Efficient Global Mapping from Sparse Label Data}},
      author={Christopher F. Brown and Michal R. Kazmierski and Valerie J. Pasquarella and William J. Rucklidge and Masha Samsikova and Chenhui Zhang and Evan Shelhamer and Estefania Lahera and Olivia Wiles and Simon Ilyushchenko and Noel Gorelick and Lihui Lydia Zhang and Sophia Alj and Emily Schechter and Sean Askay and Oliver Guinan and Rebecca Moore and Alexis Boukouvalas and Pushmeet Kohli},
      year={2025},
      eprint={2507.22291},
      archivePrefix={arXiv},
      primaryClass={cs.CV},
      note={arXiv:2507.22291}
}

@misc{alain2016understanding,
      title={{Understanding intermediate layers using linear classifier probes}}, 
      author={Guillaume Alain and Yoshua Bengio},
      year={2018},
      eprint={1610.01644},
      archivePrefix={arXiv},
      primaryClass={stat.ML},
      note={arXiv:1610.01644}, 
}

@article{belinkov2022probing,
  title   = {Probing Classifiers: Promises, Shortcomings, and Advances},
  author  = {Belinkov, Yonatan},
  journal = {Computational Linguistics},
  volume  = {48},
  number  = {1},
  pages   = {207--219},
  year    = {2022}
}

@inproceedings{blumenstiel2025terramesh,
author    = {Blumenstiel, Benedikt and Fraccaro, Paolo and Marsocci, Valerio and Jakubik, Johannes and Maurogiovanni, Stefano and Czerkawski, Mikolaj and Sedona, Rocco and Cavallaro, Gabriele and Brunschwiler, Thomas and Moreno, Juan Bernabe and Longepe, Nicolas},
title     = {{TerraMesh}: A Planetary Mosaic of Multimodal {Earth Observation} Data},
booktitle = {Proceedings of the IEEE/CVF Conference on Computer Vision and Pattern Recognition Workshops},
pages     = {2419--2427},
year      = {2025},
month     = jun
}
\bibliographystyle{spiebib}

\end{document}